\documentclass[10pt]{fcs}

\usepackage{bm}
\usepackage{graphicx}
\usepackage{amsmath}
\usepackage{amsthm}
\usepackage{booktabs}
\usepackage{algorithm}
\usepackage{algorithmic}
\usepackage{amssymb}
\usepackage{multirow}
\usepackage{subfigure}
\usepackage{pifont}
\usepackage{makecell}
\usepackage[normalem]{ulem}
\useunder{\uline}{\ul}{}

\usepackage[T1]{fontenc}   

\title{Improved Paraphrase Generation via Controllable Latent Diffusion}
\shorttitle{Latent Diffusion Paraphraser}
\author[1]{Wei Zou}
\author[1,2]{Ziyuan Zhuang}
\author[1]{Xiang Geng}
\author*[1]{Shujian Huang}
\author[2]{Jia Liu}
\author[1]{Jiajun Chen}
\address[1]{National Key Laboratory for Novel Software Technology, Nanjing University, 210023, China}
\address[2]{Collaborative Innovation Center of Novel Software Technology and Industrialization, Nanjing University, 210023, China}
\fcssetup{
  received       = {June 24th, 2024},
  accepted       = {January 4th, 2025},
  corr-email     = {huangsj@nju.edu.cn},
}

\begin{abstract}
Paraphrase generation strives to generate high-quality and diverse expressions of a given text, a domain where diffusion models excel.
Though SOTA diffusion generation reconciles generation quality and diversity, textual diffusion suffers from a truncation issue that hinders efficiency and quality control.
In this work, we propose \textit{L}atent \textit{D}iffusion \textit{P}araphraser~(LDP), a novel paraphrase generation by modeling a controllable diffusion process given a learned latent space.
LDP achieves superior generation efficiency compared to its diffusion counterparts.
It can facilitate only input segments to ensure paraphrase semantics, improving the results without external features.
Experiments show that LDP better reconciles paraphrase generation quality and diversity than baselines.
Further analysis shows that our method is also helpful to other similar text generations and domain adaptations

\end{abstract}

\begin{document}
\section{Introduction}
Paraphrase generation aims to produce semantically equivalent sentences in varied linguistic forms. 
It's versatile in generation tasks such as text summarization~\cite{zhou2021paraphrase, dong2017learning} and question answering~\cite{buck2017ask}. 
It also contributes to popular techniques such as data augmentations~\cite{kumar2019submodular}.

Despite viable mainstream paradigms, traditional paraphrasing struggles to balance generation quality and diversity.
The deterministic paradigm via the encoder-decoder model~\cite[enc-dec]{vaswani2017attention} focuses on high-quality generation instead of diversity.
Such paradigm is further improved in diversity by enforcing explicit external features of high-level semantics shared by paraphrases, such as syntax~\cite{sun2021aesop} and examplars~\cite{chen2019controllable}.
However, external features for diversity are not always available.
On the other hand, the variational paradigm promotes diversity via variational autoencoders, which model the latent distribution shared by diverse contextual representations~\cite{bao2019generating,du2022diverse}.
However, its sampling nature given condensed Gaussian distributions hinders the generation quality of complex language patterns despite additional keyword guidance~\cite{chen2022mcpg}. 

Images and speech recently witnessed a novel generation paradigm via the diffusion probabilistic model~\cite[DPM]{ho2020denoising} achieving SOTA in both quality and diversity.
The DPM is the generation that morphs toward high-quality data through numerous rounds of continuous Markov transitions.
Though DPM shares similar stochasticity with the variational generation, it does not fall short in quality.
Additionally, its neat interventions enable the continuous diffusion transitions to meet the quality required by versatile data generations~\cite{nichol2021improved}, which is prominent in diffusion implementation.
However, DPM works in a continuous space, which differs from the discrete text token.

Lately, Diffusion-LM~\cite{DiffusionLM2022Lisa} and D3PM~\cite{austin2021structured} cater to text generation via an additional discrete sampling called `rounding' process.
The `rounding' essentially bridges the continuous diffusion representations with corresponding discrete tokens, as arbitrary diffusion intervals are truncated to embeddings of valid texts with additional decodings.
However, `rounding' introduces decoding overhead, thus hindering high-efficiency generation by SOTA diffusion implementations~\cite{ye2023dinoiser,karras2022elucidating}.
Furthermore, `rounding' also introduces truncation errors when arbitrary diffusion intervals are rounded to specific token embeddings, further hindering possible intervention.
Therefore, we consider circumventing the truncation issue to improve efficiency and enable generation interventions when paraphrasing via text diffusion.

In this work, we propose the \textit{L}atent \textit{D}iffusion \textit{P}araphraser (LDP), which not only enjoys the benefit of DPM in balancing quality and diversity but also utilizes flexible control in diffusion modeling for paraphrasing.
LDP adopts the latent space from a given enc-dec framework, which offers more efficacy than raw features for diffusion as suggested by work~\cite{rombach2022high,lovelace2022latent}.
The off-the-shelf encoder and decoder bridge the continuous diffusion process with corresponding discrete texts. 
Thus, LDP prevents the intermediate roundings required by diffusion on raw text, which offers generation efficiency.
Furthermore, removing roundings enables SOTA diffusion controls, where we can even utilize only input segments rather than external features to ensure semantics for improvement.
Experiments show that LDP achieves better and faster paraphrase generation than its diffusion counterparts on various datasets.
Further analysis shows that our methods are viable to other similar text generations and domain adaptation.

Our contributions can be summarized as follows:
\begin{itemize}
    \item We propose a novel paraphrase generation called LDP, which better reconciles generation quality and diversity elegantly and efficiently. 
    \item LDP utilizes the flexible diffusion control on semantics for improvement even with only input segments rather than external features.
    \item Analysis shows that our method is also helpful in similar text scenarios such as question generation and domain adaptation.
\end{itemize}
\section{Preliminary}

\subsection{Paraphrase Generation}

Paraphrase refers to the diverse utterances that keep the original semantic.
Paraphrase generation is crucial in several downstream natural language processing~(NLP) tasks.
Though methods based on deterministic seq2seq framework achieve success~\cite{vaswani2017attention,sancheti2022entailment,yang2019end}, the nature of maximum likelihood estimation hinders the generation diversity. 
Some researchers promote generation diversity by enforcing explicit external features of high-level semantics shared by paraphrases such as syntactic structures or exemplar syntax~\cite{hosking2022hierarchical, yang2022gcpg}, which are not easily accessible. 
Others turn to variational generation to fit the shared latent distribution of diverse textual representations~\cite{bowman2015generating,du2022diverse}.
However, variational generations sacrifice the quality for its diversity.

\subsection{Latent Diffusion Models with Control}

The diffusion probabilistic model~\cite[DPM]{ho2020denoising} is a Markov chain of variational reconstruction of the original inputs $z_{0}$ given data distribution $q(z)$ from Gaussian-distributed noise.
Specifically, the DPM is trained by sampling from a Markov noising process $P(z_{t+1}|z_{t}) \sim \mathcal{N}(z_{t+1};\sqrt{1-\beta_t}z_{t}, \beta_t \mathcal{I})$ scheduled by series of noise scales $\beta_t\in(0,1)$, where the $\beta_t$ is considered the standard deviation of the step-wise transition distribution $\sqrt{\beta_t}=\sigma_t$ at step $t\in[0, T]$.
Work~\cite{rombach2022high} further proposes the latent diffusion where the diffusion process is introduced to the latent space of a well-trained enc-dec framework.
The latent diffusion achieves better results for video~\cite{he2023latent}, audio~\cite{liu2023audioldm}, and image synthesis~\cite{lai2023minidalle3} since diffusion upon encoded latent features proves more efficacy than that upon raw representations.

Training a diffusion model follows the intuition to fit reconstruction from noised $z_t$ to its original $z_0$ along a T-step Markov-Gaussian noising process:
\begin{equation}
    \mathcal{L} = \mathbb{E}_{t\sim [0, T]}\left[\|z_\theta(z_t, t)-z_0\|^2_2\right],
    \label{eq:train}
\end{equation}
where $z_\theta(\cdot)$ is a reconstruction neural net given noised $z_t$ and noising step $t$.
Work~\cite{ho2020denoising} further deducted t-steps Markov-Gaussian noising process into a one-step noising, leading to a closed form $z_t$ by given noising step $t$:
\begin{equation}
    z_t = \sqrt{\overline{\alpha}_t}z_0 + \sqrt{1-\overline{\alpha}_t}\epsilon, \epsilon \sim \mathcal{N}(0, \mathcal{I}),
    \label{eq:forward}
\end{equation}
where $\overline{\alpha}_t=\Pi^t_{i=1}(1-\beta_i)$, as $t$ determines the noise schedule $\beta_t$.

The corresponding diffusion generation iteratively morphs from pure Gaussian noise to valid data by the learned reconstructor $z_\theta(\cdot)$.
That is, given a fully optimized $z_\theta(\cdot)$, the generation starts from pure Gaussian sample $z_T \in \mathbb{R}^{l \times d} \sim \mathcal{N}(0, I)$ with a roughly reconstructed $\hat{z}_0$, then iteratively noise and denoise by diminishing noise scales determined by $\beta_t$.
Such iteration can be conducted via various diffusion sampling algorithms such as ancestral sampling~\cite{ho2020denoising}, DDIM sampler~\cite{song2020denoising}, or ODE solvers\cite{lu2022dpm,lu2022dpmp}.

\begin{figure*}[ht]
    \centering
    \includegraphics[width=0.8\linewidth]{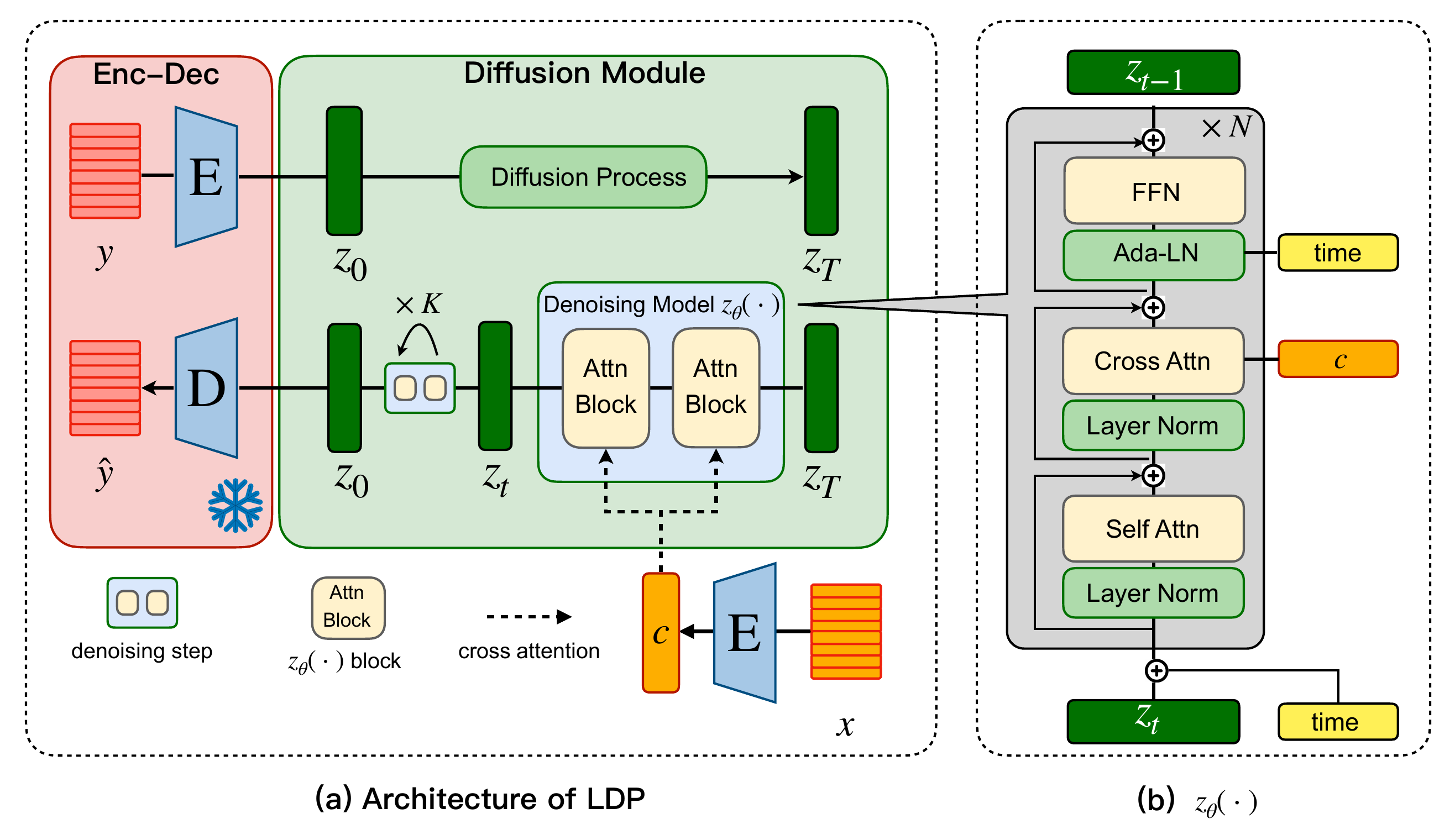}
    \caption{Overview of LDP.
    (a) Model architecture. LDP consists of an encoder-decoder~(enc-dec) framework~(pink) and a diffusion module~(green). 
    The encoder~(E) and decoder~(D) are frozen to bridge the continuous diffusion process with corresponding discrete texts.
    Where the source $x$ is encoded as $c$ during the diffusion on latent variable $z$. 
    (b) Detailed architecture of denoising model $z_\theta(\cdot)$. 
    }
    \label{fig:overview}
\end{figure*}

Plug-and-play modules can flexibly intervene in iterative diffusion generation with minor overheads, where the SOTA model is ControlNet~\cite{zhang2023adding}.
ControlNet is an adaptor that finetunes a trainable copy of the DPM $z_\theta'(\cdot)$ to cater to versatile generations while freezing the learned DPM $z_\theta(\cdot)$ to maintain harmless adaptation. 
The trainable copy $z_\theta'(\cdot)$ takes in additional control signals to the DPM via zero-convolution layers, i.e., convolution layers initialized by zero weight and bias.
The controller inputs $\delta$ are fused with original diffusion steps by Eq~\ref{eq:controlnet}:
\begin{equation}
     \begin{split}
    {z}_{t+1} =& z_\theta(z_t, t) + \text{zero\_conv}(z_\theta'(z_t+\text{zero\_conv}(\delta), t)),
    \end{split}
    \label{eq:controlnet}
\end{equation}
where $\delta$ enables vague yet versatile guidance such as Canny edges and poses~\cite{zhang2023adding}.

\section{Methodology}
In this section, we detail \textit{L}atent \textit{D}iffusion \textit{P}araphraser (LDP).
LDP models the diffusion process upon the latent space of a pretrained enc-dec framework, where mere input segments can further guide paraphrase semantics for improvement.

\subsection{Latent Diffusion Paraphraser}
The overall pipeline is shown in Fig~\ref{fig:overview}(a).
LDP consists of an enc-dec framework to provide a learned latent space and a diffusion module.
The encoder and decoder with a learned latent space bridge the continuous diffusion process with corresponding discrete texts once and for all at the beginning and the end of the generation, instead of step-wise rounding.
Notably, LDP is compatible with the mainstream enc-dec framework as long as it bijectively maps the texts with the corresponding latent representations.

To make life easier, we adopt BART~\cite{lewis2019bart} for illustration, which is an off-the-shelf pretrained language model that encodes and decodes arbitrary text with its corresponding latent representation.
Given a text sequence $x = \{x_1, x_2, \dots, x_l\}$, the BART encoder $\text{E}$ encodes it into a $d$-dimentional latent representation $z=\text{E}(x),z\in\mathbb{R}^{l\times d}$ for diffusion process, while the decoder $\text{D}$ yields corresponding text sequence given the latent representation $y\thickapprox\text{D}(z) = \text{D}(\text{E}(x))$.
The model utilizes T5 relative positional embeddings~\cite{raffel2020exploring}.
Input features for diffusion are normalized by training data features, where we adopt BART encodings for mean and standard deviation~\cite{rombach2022high}.
Concordantly, the normalization is reversed before text decoding.
The encoder and decoder are frozen during diffusion training, thus leaving the reconstruction network $z_\theta(\cdot)$ the only trainable parameters.

Given paraphrase pair $\langle x,y \rangle$, we then train the end-to-end diffusion models for encoder latent space, where the reconstruction network is parameterized by $z_\theta(z_t,c,t)$ with addition source encoding $c=E(x)$ as input.
$z_\theta(\cdot)$ consists of $N$ layers of pre-layer norm transformer blocks. 
As shown in Fig~\ref{fig:overview}(b), each layer consists of a self-attention encoding for $z_t$, followed by a cross-attention access of the encoded source $c$ and a time step interpolation via a feedforward layer.
The time step $t$ is embedded as a d-dimensional vector, then its interpolation is preprocessed by AdaLN~\cite[adaptive layer norm]{DBLP,peebles2022scalable} instead of the generic layer norm.
AdaLN regresses the layer-wise normalization scale and shifts from the sum of time embedding and encoded input features.
The network layers are activated by GeGLU~\cite{shazeer2020glu} following SOTA transformer implementation~\cite{raffel2020exploring}.

$z_\theta(\cdot)$ is optimized by reconstruction loss in Eq~\ref{eq:train}.
We uniformly sample time step $t$ for arbitrary noised representation $z_t$ by Eq~\ref{eq:forward}, where $t$ determines the noise scale $\beta_t$ by noise schedule~\cite{ho2020denoising,nichol2021improved}.
We also apply sentence-level condition dropout during training to ensure the model's unconditional language generation, that is, to replace the source sentence representation with trainable null tokens $y_\varnothing$ with probability $p=0.1$.

The optimized $z_\theta(\cdot)$ is implemented in SOTA diffusion samplers~\cite{song2020score,lu2022dpm,lu2022dpmp}, which will morph pure Gaussian noise $z_T$ into the latent representation $\hat{z}_0$ of the given source text.
$D$ further decodes $\hat{z}_0$ for text output.
Unlike generation with length prediction, LDP determines the sequence length by end-of-sequence label automatically.


\subsection{Semantics Guidance by Controller}

\begin{figure}[ht]
    \centering
    \includegraphics[width=0.8\linewidth]{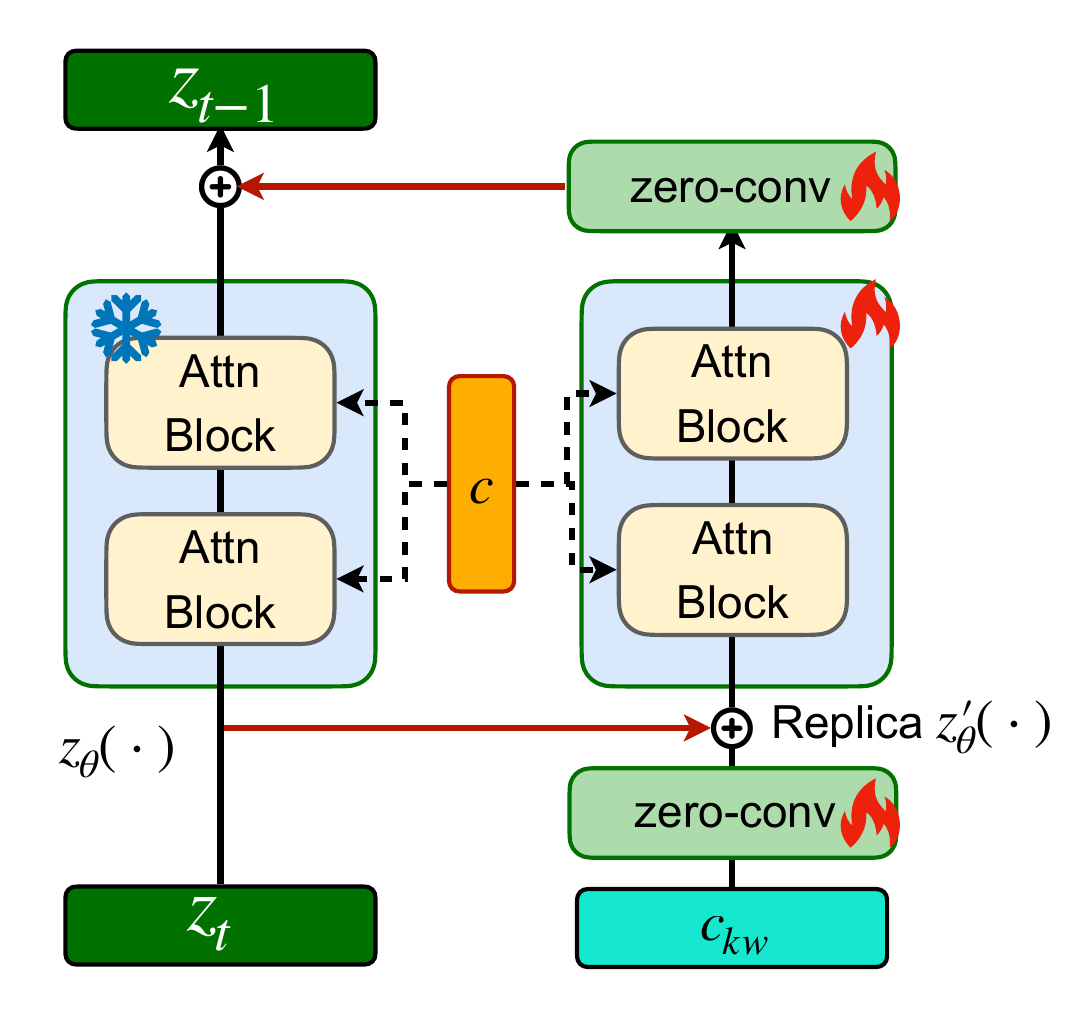}
    \caption{Architecture of LDP controller, where $c_{kw}$ indicates the encoded keyword segments.
    We freeze the learned $z_\theta(\cdot)$, then finetune the replica $z'_\theta(\cdot)$}
    \label{fig:controlnet}
\end{figure}

Diffusion generation can introduce additional controls via ControlNet~\cite{zhang2023adding}, which is a step-wise plug-and-play adaptor framework.
It has proved its efficacy by manipulating image generation given vague constraints such as canny edges or skeleton poses.
By removing step-wise rounding, LDP circumvents the truncation issue against control signals, where we consider paraphrase generation can harness input segments as the vague constraints likewise.

Intuitively, the paraphrase can be guided by such a framework based on vital semantic fragments for better generation.
We leverage input tokens above a certain length as keywords, guiding the diffusion generation on given semantics to improve paraphrase.
Our implementation is shown in Fig~\ref{fig:controlnet}.
We sample from the longest $15\%$ tokens in a given sentence as keywords and mask the remaining parts with placeholder~\verb|<M>| as semantic segments. 
The semantic segments are then encoded by the same encoder used by the original DPM for controller inputs.
The controller block $z_\theta'(\cdot)$ is a trainable copy of the well-trained $z_\theta(\cdot)$, while $z_\theta(\cdot)$ is frozen.
The controller inputs are fused with original diffusion generation by Eq~\ref{eq:controlnet} as:
\begin{equation*}
     \begin{split}
    \hat{z}_0 =& z_\theta(z_t, c, t) + \text{zero\_conv}(z_\theta'(z_t+\text{zero\_conv}(c_{kw}), c, t)),
    \end{split}
\end{equation*}
where $c_{kw}$ refers to the encoded keyword segment.
During fine-tuning, we extract keyword segments from reference paraphrases, while for inference, we likewise harness the source input for keyword segments.

\begin{table*}[ht]
\centering
\resizebox{\textwidth}{!}{
\begin{tabular}{l|llll|llll|llll}
\hline
                     & \multicolumn{4}{c|}{QQP}                                                                     & \multicolumn{4}{c|}{WIKI}                                                                    & \multicolumn{4}{c}{Twitter}                                                                   \\
                     & BLEU  & PPL    & div-4 & \begin{tabular}[c]{@{}l@{}}iBScore \\ (BScore-srcBLEU)\end{tabular} & BLEU  & PPL    & div-4 & \begin{tabular}[c]{@{}l@{}}iBScore \\ (BScore-srcBLEU)\end{tabular} & BLEU  & PPL     & div-4 & \begin{tabular}[c]{@{}l@{}}iBScore \\ (BScore-srcBLEU)\end{tabular} \\ \hline
Transformer          & 30.36 & 214.98 & 58.01 & \begin{tabular}[c]{@{}l@{}}33.53 \\ (86.92-55.39)\end{tabular}      & 50.64 & 202.60 & 68.32 & \begin{tabular}[c]{@{}l@{}}13.15 \\ (89.91-76.76)\end{tabular}      & 31.06 & 299.47  & 58.42 & \begin{tabular}[c]{@{}l@{}}48.11 \\ (81.90-33.79)\end{tabular}      \\
BART-FT              & 33.73 & \textbf{189.23} & 57.20 & \begin{tabular}[c]{@{}l@{}}33.94 \\ (54.32-88.26)\end{tabular}      & \textbf{62.79} & 112.51 & 61.67 & \begin{tabular}[c]{@{}l@{}}10.87 \\ (93.89-83.02)\end{tabular}      & \textbf{33.21} & 324.14  & 58.15 & \begin{tabular}[c]{@{}l@{}}47.10 \\ (81.74-34.64)\end{tabular}      \\
T5-GPVAE             & 33.50 & 200.13 & 64.15 & \begin{tabular}[c]{@{}l@{}}25.89 \\ (87.84-61.95)\end{tabular}      & 61.05 & \textbf{109.07} & 63.98 & \begin{tabular}[c]{@{}l@{}}10.89 \\ (89.91-76.76)\end{tabular}      & 27.01 & \textbf{129.86}  & 58.59 & \begin{tabular}[c]{@{}l@{}}52.82 \\ (79.83-27.01)\end{tabular}      \\
BART-CVAE            & 32.21 & \underline{194.81} & 55.40 & \begin{tabular}[c]{@{}l@{}}47.02 \\ (87.56-40.54)\end{tabular}      & 61.14 & \underline{116.88} & 64.16 & \begin{tabular}[c]{@{}l@{}}9.68 \\ (93.81-84.13)\end{tabular}       & \underline{32.24} & 325.71  & 59.77 & \begin{tabular}[c]{@{}l@{}}44.37 \\ (81.48-37.11)\end{tabular}      \\
\hline
DiffuSeq             & 31.49 & 397.68 & 86.41 & \begin{tabular}[c]{@{}l@{}}49.11 \\ (85.89-36.78)\end{tabular}      & 25.89 & 440.49 & \underline{80.67} & \begin{tabular}[c]{@{}l@{}}48.35 \\ (84.47-36.12)\end{tabular}      & 9.83  & 1045.88 & 85.45 & \begin{tabular}[c]{@{}l@{}}54.61 \\ (75.06-20.46)\end{tabular}      \\
SeqDiffuSeq          & 26.67 & 404.28 & \textbf{70.35} & \begin{tabular}[c]{@{}l@{}}48.28 \\ (84.56-36.28)\end{tabular}      & 28.13 & 309.88 & 73.13 & \begin{tabular}[c]{@{}l@{}}13.15 \\ (89.91-76.76)\end{tabular}      & 11.92 & 903.26  & 71.51 & \begin{tabular}[c]{@{}l@{}}52.46 \\ (76.62-24.16)\end{tabular}      \\
\hline
Llama2-7b            & 16.93 & 218.02 & 66.22 & \begin{tabular}[c]{@{}l@{}}50.10 \\ (85.95-35.85)\end{tabular}      & 23.26 & 166.70 & 62.09 & \begin{tabular}[c]{@{}l@{}}60.73 \\ (86.74-26.01)\end{tabular}      & 4.28  & 262.46  & 60.65 & \begin{tabular}[c]{@{}l@{}}57.47 \\ (73.74-16.27)\end{tabular}      \\
Llama3.1-8b          & 13.90 & 210.21 & 67.23 & \begin{tabular}[c]{@{}l@{}}\textbf{57.97} \\ (85.37-27.40)\end{tabular}      & 16.56 & 187.29 & 60.67 & \begin{tabular}[c]{@{}l@{}}\underline{63.31} \\ (85.35-23.03)\end{tabular}      & 3.42  & \underline{240.44}  & 57.87 & \begin{tabular}[c]{@{}l@{}}\textbf{61.48} \\ (75.25-13.77)\end{tabular}      \\ \hline
LDP(ours)            & \underline{36.56} & 267.53 & 73.22 & \begin{tabular}[c]{@{}l@{}}50.57 \\ (87.09-36.52)\end{tabular}      & 60.38 & 120.13 & \textbf{81.2}  & \begin{tabular}[c]{@{}l@{}}58.15 \\ (85.15-27.00)\end{tabular}      & 18.75 & 466.46  & \textbf{93.32} & \begin{tabular}[c]{@{}l@{}}59.72 \\ (76.14-16.42)\end{tabular}      \\
\ w/ SG & \textbf{37.48} & 246.76 & \underline{73.63} & \begin{tabular}[c]{@{}l@{}}\underline{51.44} \\ (88.82-37.40)\end{tabular}      & \underline{61.99} & 146.54 & 72.22 & \begin{tabular}[c]{@{}l@{}}\textbf{64.91} \\ (91.34-26.76)\end{tabular}      & 19.72 & 419.63  & \underline{93.26} & \begin{tabular}[c]{@{}l@{}}\underline{60.36} \\ (82.05-21.69)\end{tabular}      \\ \hline
\end{tabular}
}
\caption{Results on paraphrase generation. 
Baselines are implemented from the source code. 
The bold indicates the top results, whereas the second bests are underlined.
We group the SOTA baselines into two groups, the traditional and diffusion baselines. 
LDP outperforms SOTA diffusion baselines of traditional baselines.
Notably, open-source LLMs have a model size of 30-40$\times$ larger than LDP.
LDP with semantics guidance~(LDP w/SG) evan achieves the iBScore equivalent to open-source LLMs, with significantly higher BLEU scores. 
}
\label{tab:main-result}
\end{table*}

\section{Experiment}
\subsection{Setups}
\paragraph*{Implementation Details}

We implemented LDP by N=12 layers of Transformer blocks with 12 attention heads, where we adopt time step embedding in the same way as the position embedding.
The maximum sequence length is $96$.
We adopt \textit{BART-base}~(139M) for encoder and decoder, and the diffusion dimension is $768$, the same as BART embedding dimension.

The diffusion process is defined by $T=1000$, with cosine schedule~\cite{nichol2021improved} for $\beta_t$.
We train the model for up to $250$k steps with batch size $192$.
The learning rate for the DPM training is $10^{-4}$, while for the controller fine-tuning is $10^{-5}$.
We apply DPM-Solver++\cite{lu2022dpmp} with 25 steps for diffusion sampling.


We trained our model on 4$\times$ v100 and 4$\times$ Nvidia 3090 GPUs for about 100 GPU-hours.
LDP has approximately 200M total parameters.

\paragraph*{Datasets}
We adopt Quora Question Pairs (QQP), Twitter-URL (Twitter)~\cite{lan2017continuously} and positive pairs from PAWS-wiki~\cite{paws2019naacl}, which is popular amongst mainstream paraphrase generations.
For QQP and Twitter data, we randomly divided both datasets into three parts: 10k test with 10k validation sentences, and the remaining were assigned to the training set.

\paragraph*{Baselines}
We first adopt several mainstream paraphrase generations as baselines, including the deterministic paradigm by generic Transformer~\cite{vaswani2017attention}, fine-tuned BART-base~(BART-FT)~\cite{lewis2019bart}, and variational paradigm equipped with pretrained encoder and decoder by T5-GPVAE~\cite{du2022diverse} and BART-CVAE~\cite{wang2019t}.
We also include SOTA text generations via the diffusion model, such as DiffuSeq~\cite{gong2022diffuseq} and SeqDiffuSeq~\cite{yuan2022seqdiffuseq}, which are versatile end-to-end diffusion modeling.
DiffuSeq applies minimum Bayes risk decoding~\cite[MBR]{kumar2004minimum} with 10 candidates, while SeqDiffuSeq implements beam search for decoding given an on-the-fly enc-dec framework during rounding.
For all beam search in the experiments, we apply beam size as $4$.
Last but not least, we adopt the latest open-source LLMs for comparison, Llama2-7b~\cite{touvron2023llama2} and Llama3.1-8b~\cite{dubey2024llama3}\footnote{Following reviewers' recommendations, we incorporate LLMs into our comparison, which were introduced a year after our initial draft.}, which are supervised fine-tuned via LoRA~\cite{hu2021lora} given the same training data on the same device.
We adopt $rank$ and $lora\_\alpha$ by $(64,128)$ for one-epoch LoRA tuning.

\paragraph*{Metrics} 
\begin{table*}[ht]
\centering
\begin{tabular}{cl}
\hline
src           & how is black money gon na go off with no longer the use of the same 500 and 1000 notes ? \\
original paraphrase        & how does black money brought out to black money market or corruption? \\
enforcement   & \verb|<M>| \verb|<M>| black money \verb|<M>| \verb|<M>| \verb|<M>| \verb|<M>| \verb|<M>| no longer \verb|<M>| \verb|<M>| \verb|<M>| \verb|<M>| \verb|<M>| \verb|<M>| 1000 \verb|<M>| \verb|<M>| \\
Guided paraphrase & how does banning 500 and 1000 rupee notes solve the black money problem? \\
\hline
\end{tabular}
\caption{
Guide paraphrase semantics by input segments.
One can highlight the input segments to guide the paraphrase.
We inject input segments masked by placeholder <M> of `black money', `no longer', and `1000' via controller, which improves the paraphrase.
}
\label{tab:enforce_case}
\end{table*}

We first cater to the traditional paraphrase evaluation for generation quality with reference-oriented BLEU~\cite{papineni2002bleu} and perplexity~(PPL) based on open-source GPT2-base model~\cite{radford2019language};
However, an ideal paraphrase must reconcile both generation quality and diversity. 
We adopt the distinct n-gram~\cite[div-4]{deshpande2019fast} for diversity, which measures the ratio of unique n-gram tokens to the total generations.
Considering the stochastic nature of variational and diffusion generation, we have gathered the top-5 beam search samples for deterministic baselines (including fine-tuned LLMs), as well as 5 random samples for both the variational baselines and our proposed method with or without the controller.
Quality metrics are averaged for fair comparison.

Note that reference-oriented metrics contradict the evaluation of generation diversity, where the paraphrase can sacrifice either quality for diversity via nonsense generation, or diversity via duplication.
Therefore, we adopt the iBScore~\cite{dou2022improving} as the \textit{primary metric}, which measures the semantic via embedding similarity, namely the BERTScore~\cite{bert-score}, and punishes source duplication by source BLEU:
$$\text{iBScore} = \text{BERTScore} - \text{srcBLEU},$$
where we calculate BERTScore via deBERTa-large~\cite{he2021deberta}.

Lastly, we  additionally refer to the human evaluation~(HE) in work~\cite{shen2022evaluation} and sample 100 paraphrases from each test set for human evaluation ranging from $0$ to $6$ by the following criterion:
\begin{itemize}
    \item 0: The paraphrase is nonsensical, failing to convey coherent meaning or with little expressive changes.
    \item 1: The paraphrase is nonsensical, failing to convey coherent meaning or with unfluent expression.
    \item 2: There is a certain amount of expression changes in the paraphrase, or significant changes that partially preserve the meaning of the source text, albeit with some unfluency or omissions.
    \item 3: There is a certain amount of expression changes in the paraphrase \textit{with better fluency}, albeit some inaccuracies.
    \item 4: There are expressive changes in paraphrase, but it largely retains its semantics with minor issues such as slight grammar errors.
    \item 5: There are expressive changes in paraphrase while retaining the semantics with fluency.
    \item 6: There are creative or exceeding expressions in paraphrases while accurately conveying the full meaning of the source text without errors.
\end{itemize}
The scores support the half value in between.
The sampled paraphrases are double-blind for random assignment, each with at least two scorers in case of bias. 
The above metrics are collected on generated samples and averaged for the final results.  
\begin{table}[ht]
\centering
\begin{tabular}{c|l}
                     & HE    \\
\hline
Transformer-base          & 3.9   \\
BART-FT              & 4.465 \\
T5-GPVAE             & 4.355 \\
BART-CVAE            & 4.255 \\
DiffuSeq             & 3.655 \\
SeqDiffuSeq          & 3.32  \\
\hline
Llama2-7b            & 4.63  \\
Llama3.1-8b          & \textbf{4.915} \\
\hline
LDP(ours)            & 4.585 \\
   w/ SG             & \underline{4.75} 
\end{tabular}
\caption{
    Results of human evaluation~(HE).
    The top result is bolded, and the second best is underlined.
    LDP exceeds traditional and diffusion baselines. 
    Compared to the SOTA LLMs, LDP achieves comparable results with the adaptor for semantics guidance~(with SG).
    }
\label{tab:HE}
\end{table}

\begin{table}[ht]
\resizebox{0.45\textwidth}{!}{
\begin{tabular}{cl}
\hline
Source                    & what should i do to improve my tennis? \\
\hline
\multirow{5}{*}{DiffuSeq} & \makecell[l]{what is the best way to improve your tennis\\ \textbf{andtiv} month?} \\
                          & what should i do to improve my tennis?                    \\
                          & how can i increase tennis?                                \\
                          & how can i improve my tennis?                              \\
                          & how do i improve my tennis skills?                    \\
\hline
\multirow{2}{*}[-4ex]{BART}   & how can i improve tennis skills?                          \\
                          & how can i get better in tennis?                           \\
\multirow{3}{*}{-CVAE}                          & how do i improve my tennis?                  \\
                          & how do i improve tennis playing?                \\
                          & how can i improve tennis skills?              \\
\hline
\multirow{5}{*}{LDP}      & what is the best way to be good at tennis?             \\
                          & what are the best ways to get better at tennis?           \\
                          & how can i improve to get better at tennis?                \\
                          & \makecell[l]{how can i improve my skills at professional\\ tennis?}       \\
                          & how can i improve my skill for tennis playing?            \\
\hline
\end{tabular}
}
\caption{
LDP generates more fluent and diverse paraphrases compared to baselines.
DiffuSeq even generates errors like `\textbf{andtiv}'.
}
\label{tab:case}
\end{table}

\subsection{Main Results}
Table~\ref{tab:main-result} presents the main results of our experiments, with human evaluation in Table~\ref{tab:HE}.
LDP achieves comparable or even superior performance compared to traditional baselines.
Moreover, it outperforms other diffusion counterparts on QQP and Twitter test sets.
LDP achieves the best performance amongst all baselines for the QQP test set.

Specifically, LDP improves overall results~(iBScore), especially diversity~(div-4) compared to traditional end-to-end paradigms such as fine-tuned BART, where we consider the diffusion module significantly improves the generation diversity.
On the other hand, LDP also outperforms SOTA diffusion baselines such as SeqDiffuSeq in terms of BLEU and perplexity, where we consider the LDP to implement a better bridge between the diffusion process and the discrete texts than rounding.
Additionally, unlike the intuition to introduce external features, the original case output improves greatly when we guide the generation semantics with a segment of source inputs as shown in Table~\ref{tab:enforce_case}.
Though fine-tuned LLM outperforms others in human evaluations, it requires significantly larger resources and a pre-training corpus with significantly weaker reference-oriented BLEU scores.
In contrast, LDP (with or without semantic guidance) achieves comparable results with substantially reduced overhead.

As shown in Table~\ref{tab:case}, we generate several paraphrases by different latent sampling for generations.
Though DiffuSeq and BART-CVAE are SOTA generators for diversity, they still yield resemble paraphrases.
LDP, on the other hand, yields better and more diverse paraphrases.

\begin{table}[ht]
\centering
\begin{tabular}{ccccccccc}
\hline
& \multicolumn{2}{c}{QQP}                                                                                                \\
Model                                  & BLEU$\uparrow$ & BERTScore$\uparrow$  \\
\hline
DiffuSIA\cite{tan2023diffusia}         & 24.95          & 83.62             \\ 
BG-DiffuSeq\cite{tang2023can}          & 26.27          & -                  \\
TESS\cite{mahabadi2023tess}            & {\ul 30.2}     & {\ul 85.7}          \\
Dinoiser\cite{ye2023dinoiser}          & 26.07          & -                  \\
Diff-Glat\cite{qian2022diff}           & 29.86          & -                  \\
SeqDiffuSeq\cite{yuan2022seqdiffuseq}  & 24.32          & -                  \\
DiffuSeq\cite{gong2022diffuseq}        & 24.13          & 83.65               \\
LDP(ours)                              & \textbf{36.56} & \textbf{87.51}      \\
\hline
\end{tabular}
\caption{Results on QQP dataset compared with more diffusion generation baselines}
\label{tab:diffusion-qqp}
\end{table}

We additionally include more recent text diffusion generators for comparison for QQP tests.
Table~\ref{tab:diffusion-qqp} shows that LDP achieves SOTA BLEU and BERTScore amongst the diffusion baselines.

Overall, LDP with semantic controller achieves the best iBScore amongst traditional and diffusion baselines, indicating a high-quality and diverse paraphrase generation.

\section{Analysis}

\subsection{Inference Efficiency}
\begin{table}[ht]
\centering
\footnotesize
    \begin{tabular}{cccc}
    \hline
                          & Lapse~(s)                 \\
    \hline
    Transformer           & 235  (206.3$\times$)          \\
    BART-FT               & 238  (203.7$\times$)          \\
    T5-GPVAE              & 3909  (12.4$\times$)           \\
    BART-CVAE             & 252   (192.4$\times$)          \\
    \hline
    DiffuSeq(DDIM 2000)   & 48480  (1$\times$)              \\
    DiffuSeq(DDIM 500)    & 11530  (4.2$\times$)            \\
    SeqDiffuSeq(DDIM 2000)& 13851  (3.5$\times$)            \\
    LDP(ours)             & 290   (\textbf{167.2$\times$}) \\
    \hline
    \end{tabular}
    \caption{Inference overhead~(seconds) with acceleration rate~(in brackets) on the QQP validation set, where we set the DiffuSeq as the efficiency baseline.}
    \label{tab:inference-efficiency}
\end{table}

By eliminating the rounding process in text diffusion models, LDP achieves improved efficiency.
We compare several baseline efficiencies under their best-generation performance.
As shown in Table~\ref{tab:inference-efficiency}, LDP is $167.2$ times faster than DiffuSeq and $50$ times faster than SeqDiffuSeq, where the diffusion overhead takes up only $18\%$ of the total timelapse.
Consequently, our approach achieves a similar efficiency compared to that of the autoregressive baselines with only enc-dec overheads.

Note that the rounding takes up considerable computation resources for minimum Bayes risk decoding, which limits the maximum dimension supported by diffusion.
Therefore, DiffuSeq is modeled by only $128$ dimensions.
However, LDP by $768$ dimensions is still $10.4\times$ faster than DiffuSeq in a single-step text diffusion, excluding the impact of the sampling acceleration.
Thus, the removal of the rounding, thereby eliminating the expensive quality insurance like MBR, contributes the efficiency with additional model dimensions for generation quality.

\subsection{Domain Adaptation by Controller}

\begin{table}[ht]
\centering
\begin{tabular}{ccc}
\hline
          & QQP             & ChatGPT-Aug     \\
          & BLEU$\uparrow$   & BLEU$\uparrow$  \\
\hline
origin    & 36.56           & 10.16                \\
adapted       & 36.46            & 14.65 (+4.49)         \\
\hline
\end{tabular}
\caption{Domain adaptation by controller. The controller retains the diffusion adaptation }
\label{tab:adaptation}
\end{table}

The LDP with a controller retains its adapter capabilities, making it equally suitable for domain adaptation.
We additionally adopt ChatGPT-augmented paraphrase dataset~\cite[ChatGPT-Aug]{chatgpt_paraphrases_dataset} as the novel data domain, then sample $10$k sentences as the test set.
ChatGPT-Aug dataset consists of paraphrases generated by ChatGPT from an ensembled dataset including QQP, SQuAD 2.0~\cite{2016arXiv160605250R} and CNN news~\cite{see-etal-2017-get}.

We first train the LDP on QQP data for $200$k steps, then fine-tune it with a controller for ChatGPT-Aug data for $40$k steps.
The controller fine-tuning follows the same routine where the ChatGPT-Aug data pairs are training sources and references, with additional keywords from its reference as controller inputs.
Table~\ref{tab:adaptation} shows that adaptation by the controller is viable, where the fine-tuned controller substantially improves ChatGPT-Aug test BLEU, with only minor loss for the original test domain.
Note that, the performance of the original data domain is retained by inference without the controller, which outstands from traditional data adaptation by fine-tuning.

\subsection{Ablation Study for Samplers}

\begin{table}[ht]
\footnotesize
\centering
\resizebox{0.5\textwidth}{!}{
\begin{tabular}{ccccc}
\hline
                       & \multicolumn{4}{c}{QQP}         \\
Model                  & BLEU$\uparrow$ & PPL$\downarrow$ & div-4$\uparrow$ & iBScore$\uparrow$   \\
\hline
\textit{DiffuSeq}(DDIM 2000)      & 24.13          & 397.68          & 86.41           & 49.11               \\
\textit{DiffuSeq}(DDIM 500)       & 0.17           & 1195.42         & 79.36           & 52.61    \\
\textit{SeqDiffuSeq}(DDIM 2000)   & 24.32          & 404.28          & 70.35           & 48.28               \\
\hline
LDP(DDIM 500)          & 35.97          & 245.26          & 74.16           & 50.35               \\
LDP(DPM-solver 25)     & 36.56          & 267.53          & 73.22           & 50.57               \\
\hline
\end{tabular}
}
\caption{
Ablation study for diffusion samplers.
}
\label{tab:sampler}
\end{table}

The diffusion sampler plays an important role during inference, thus we conduct an ablation study on the diffusion sampler for comparison.
Intuitively, the diffusion generation improves by more steps given the same sampler.
Due to the truncation errors introduced by rounding, ODE-based samplers are not apt for text diffusion with rounding, such as DiffuSeq and SeqDiffuSeq.
Thus, we additionally adopt DDIM sampler~\cite{nichol2021improved} with 500 and 200 sampling steps, which is also adopted for diffusion baselines.
Table~\ref{tab:sampler} shows that LDP still outperforms the DiffuSeq and SeqDiffuSeq given the same sampler setting. 
Notably, LDP still outperforms the baselines by fewer sampling steps, which were expected to be inferior.

\subsection{Question Generation Ability}

\begin{table}[ht]
\centering
\begin{tabular}{cccc}
\hline
& \multicolumn{2}{c}{Quasar-T}   \\
Model          & BLEU$\uparrow$ & BERTScore$\uparrow$ \\
\hline
DiffuSIA       & 17.12          & 62.19            \\
BG-DiffuSeq    & 17.53          & -                \\
TESS           & {\ul 19.50}  & 65.8                \\
SeqDiffuSeq    & 17.5           & -                \\
DiffuSeq       & 17.31          & 61.23            \\
\hline
LDP(ours)      & 18.77          & {\ul 73.23}      \\
    +ES        & \textbf{19.50} & \textbf{73.40}   \\
\hline
\end{tabular}
\caption{Results on Quasar-T compared to other diffusion model baselines}
\label{tab:diffusion-qg}
\end{table}

Considering the compatibility of LDP besides paraphrasing, we additionally validate question generation, which aims to generate the exact question by given issue descriptions.
Question generation focuses on semantics upon relatively flexible utterance, thus we evaluate the BLEU and BERTScore on Quasar-T tests~\cite{dhingra2017quasar}.

As shown in Table~\ref{tab:diffusion-qg}, LDP with enforced semantics also achieves a comparable BLEU score with SOTA, TESS~\cite{mahabadi2023tess}, with much improved BERTScore.

\subsection{Diffusion Process during Generation}

\begin{table*}[ht]
\centering
\begin{tabular}{cl}
\hline
source & \makecell[l]{What is the best way to lose weight without diet} \\
\hline
noise scale & \makecell[c]{outputs} \\
\hline
1.0      & \makecell[l]{This is the first time since 2009 that he has not been on the losing side. He has been on and off since\\ then. He is now on the winning side. He is also on the receiving end. He was on the way home from\\ the hospital. He was on his way home when he was} \\\hline
0.8      & \makecell[l]{In the meantime, I have a feeling that this is going to be an interesting experience. The fact that I have\\ the ability to do this shows that I'm not the only one. I'm the one who has the guts to do it. And I'm\\ also the one that has the will to do} \\\hline
0.6      & \makecell[l]{however, it is not the end of the world. It is the beginning of a new year. The end of a period of time....\\The end of an era. The beginning of the year. The end to the period. The start of the next phase. The\\ first phase.} \\\hline
0.4      & \makecell[l]{how can i i lose weight? how can I lose weight? how do i? what can i lose? how did i? how do I?\\ what do i eat?} \\\hline
0.2      & how can i lose weight without doing exercise or diet?? \\\hline
0.0      & how can i lose weight without doing exercise or diet? \\
\hline
\end{tabular}
\caption{The generation process tracked by diffusion noise scale. The generation improves along the diffusion, with diminishing output length from the maximum.}
\label{tab:intermediate}
\end{table*}

\begin{figure}[ht]
    \centering
    \includegraphics[width=\linewidth]{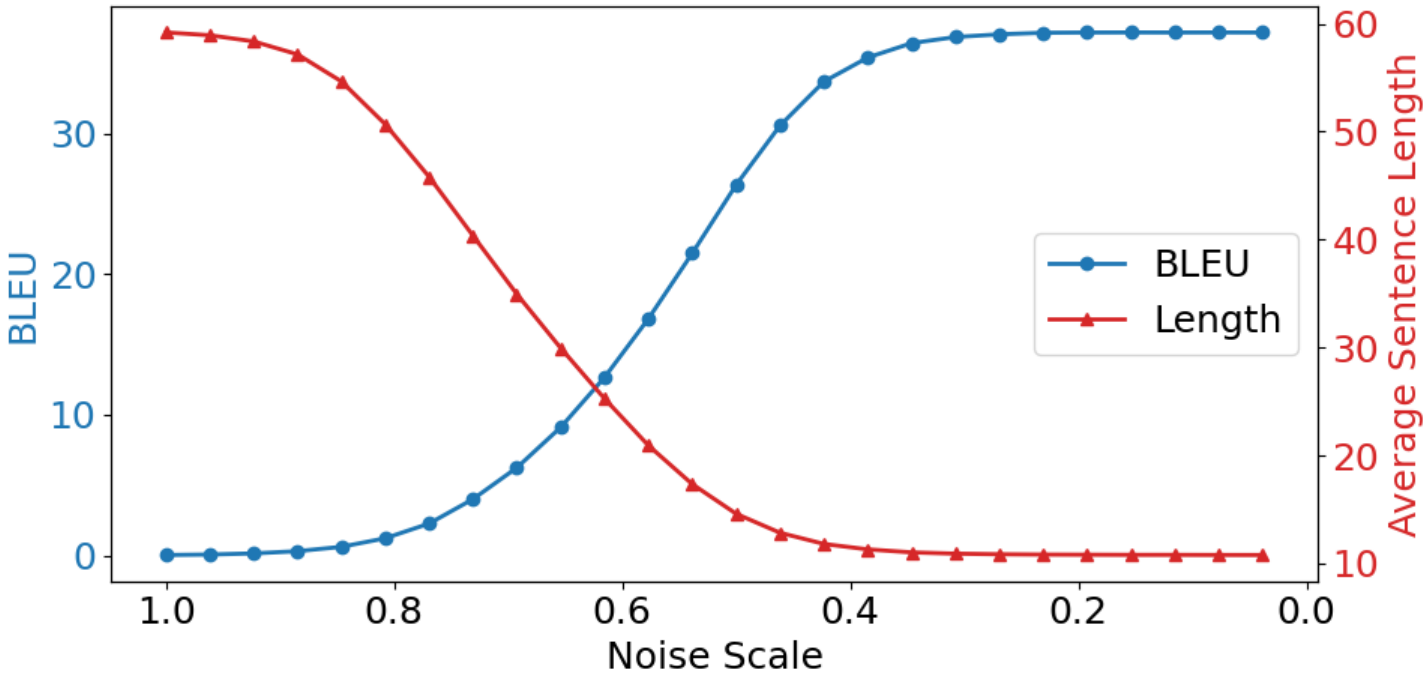}
    \caption{
    BLEU variance with averaged sentence length during LDP generation.
    The generation proceeds by diminishing noise scale with increasing BLEU and diminishing generation length.
    }
    \label{fig:inter}
\end{figure}

We track the diffusion sampling during the LDP generation by our implementation in Figure~\ref{fig:inter}, where we force-decode the intermediate latent representations as text.
Intuitively, the BLEU score increases along the sampling process, whereas the generation length will decrease from the maximum to the actual target length.
We present a generation process in Table~\ref{tab:intermediate}, where the output semantics drastically morph towards desiderata during the middle of the generation process, with only minor modifications within a smaller noise scale, as indicated by Figure~\ref{fig:inter}.

\section{Related Work}

Paraphrase generation is commonly modeled by end-to-end sequence generation, such as fine-tuned Transformers~\cite{vaswani2017attention,lewis2019bart,yang2019end}.
However, such deterministic generation fails to ensure diversity, thus some researchers~\cite{xie2022multi,sancheti2022entailment,vijayakumar2016diverse, fan2018hierarchical} introduce external features to amend.
Others turn to the variational generation for diversity~\cite{bowman2015generating}.
To amend the inferior generation by latent representation, some researchers further introduce pre-trained text encoder and decoder for generation~\cite{du2022diverse,wang2019t}.

On the other hand, the diffusion model has been a popular Markov variational generation in recent years.
Existing text diffusions cater to the discrete nature of language for versatile generations.
DiffuSeq~\cite{gong2022diffuseq} employs a single Transformer encoder and partial noising process to extend Diffusion-LM~\cite{DiffusionLM2022Lisa} for end-to-end text generations. 
They introduce discrete sampling for diffusion intervals called `rounding', to ensure generation quality.
However, rounding introduces truncation errors that hinder efficient skip-step diffusion sampler.
BG-Diffuseq~\cite{tang2023can} aims to narrow the gap between training and sampling for text diffusion via incorporating distance penalty and adaptive decay sampling.
Dinoiser~\cite{ye2023dinoiser}, turns to mitigate truncation errors by manipulating the noise scale scheduled for text diffusion training and inference to improve its efficacy.
On the other hand, SeqDiffuSeq~\cite{yuan2022seqdiffuseq} turns to a continuous diffusion modeling within an on-the-fly enc-dec framework.
Similarly, DiffuSIA~\cite{tan2023diffusia} introduces an enc-dec diffusion via spiral interaction.
Diff-GLAT~\cite{qian2022diff} incorporates residual glancing sampling, a text reconstruction via enc-dec, with its dropout rate as the noise scale.
TESS~\cite{mahabadi2023tess} proposes a logit simplex space rather than embedding space for text diffusion.

\section{Conclusion}
We propose a novel paraphrase generation via the controllable latent diffusion model, LDP, which can further incorporate flexible semantic controls to guide the generation, e.g., only input segments rather than external features.
Experiments show that LDP generates better paraphrases with superior efficiency than its diffusion counterparts.
Further analysis shows that LDP largely retains the strengths of diffusion generation and is versatile enough for similar text generation methods, such as question generation and domain adaptation.

Overall, LDP strikes a better balance between generation quality and diversity than mainstream baselines.


\begin{competinginterest}
The authors declare no competing interests. The datasets used in the training and evaluation come from publicly available sources and do not contain sensitive content such as personal information.
\end{competinginterest}

\begin{acknowledgement}
This work is supported by the National Science Foundation of China (No. 62376116, 62176120), the Liaoning Provincial Research Foundation for Basic Research (No. 2022-KF-26-02), a research project of Nanjing University-China Mobile Joint Institute.
\end{acknowledgement}

\bibliographystyle{fcs}
\bibliography{ref}

\begin{biography}{fig/zouw}
    Wei Zou is a Ph.D. candidate at the Natural Language Processing Laboratory, the National Key Lab for Novel Software Technology, Department of Computer Science \& Technology at Nanjing University, advised by Professor Shujian Huang.
    He received his B.Eng. degree from Nanjing University of Aeronautics and Astronautics, China, in 2014. 
    His research interests include natural language processing and reinforcement learning.
\end{biography}

\begin{biography}{fig/zzy}
    Ziyuan Zhuang is a master student at the National Key Lab for Novel Software Technology, Department of Software Engineering at Nanjing University, advised by Professor Shujian Huang and associate Professor Jia Liu.
    He received his B.Eng degree from Nanjing University, China in 2022.
    His research interests include natural language processing and multimodal language models.
\end{biography}

\begin{biography}{fig/gengx}
    Xiang Geng, a Ph.D. candidate at the Natural Language Processing Laboratory, the National Key Lab for Novel Software Technology, Department of Computer Science \& Technology of Nanjing University, is advised by Professor Shujian Huang. 
    His primary research interests lie in multilingual large models and the evaluation of machine translation quality. 
\end{biography}

\begin{biography}{fig/Huangsj}
    Shujian Huang is a professor at the Department of Computer Science \& Technology at Nanjing University, China. 
    He received his Ph.D. degree in the Department of Computer Science \& Technology at Nanjing University in 2012. 
    He joined Nanjing University as an assistant professor in 2018, and obtained professor in 2024. 
    His research interests majorly include natural language processing and multilingual applications.
\end{biography}

\begin{biography}{fig/Jialiu}
    Jia Liu is an associate professor at the Software Institute, Nanjing University, China.
    He received his Ph.D. degree in system engineering from Nanjing University in 2021. 
    His research interests include developing social networks, swarm intelligence, big data quality metric, and AI testing and optimization.
\end{biography}

\begin{biography}{fig/chenjj}
    Jiajun Chen is a professor at the Department of Computer Science \& Technology at Nanjing University, China. 
    His research interests include natural language processing and concurrent object-oriented programming.
\end{biography}

\end{document}